\documentclass{article}
\PassOptionsToPackage{table,xcdraw}{xcolor}

\usepackage{lineno}
\modulolinenumbers[5]

\usepackage{lipsum}
\makeatletter
\def\ps@pprintTitle{%
 \let\@oddhead\@empty
 \let\@evenhead\@empty
 \def\@oddfoot{}%
 \let\@evenfoot\@oddfoot}
\makeatother
\usepackage{amssymb}
\usepackage{amsthm}
\usepackage{amsmath}
\usepackage{cancel}
\usepackage{graphicx}
\usepackage{subcaption}
\usepackage{color}
\usepackage[usenamescdvipsnames]{xcolor}
\usepackage{float}
\usepackage{bm}
\usepackage{caption}
\usepackage{booktabs}
\usepackage{setspace} 
\usepackage[intoc]{nomencl}
\usepackage{makeidx}
\usepackage[export]{adjustbox}
\makenomenclature
\usepackage{multicol} 
\usepackage{framed}
\usepackage[hyperfootnotes=false]{hyperref}

\usepackage{t1enc}
\usepackage{srcltx}
\usepackage{amsmath}
\usepackage{nomencl}
\makenomenclature
\usepackage{amsfonts} 
\usepackage{amssymb} 
\usepackage{braket}
\usepackage[version=4]{mhchem}
\usepackage{dirtree}
\usepackage{listings}
\usepackage{graphicx}
\usepackage{grffile}
\usepackage{mhchem}

 \graphicspath{{figures/}}
 
\def\diff[#1][#2]{\frac{\partial #1}{\partial #2}}
\def\difft[#1]{\partial_t #1}

  \RequirePackage{ifthen}
  \renewcommand{\nomgroup}[1]{%
   \medskip%
              \ifthenelse{\equal{#1}{B}}%
    {\item []\hspace*{-\leftmargin}%
       \emph{Latin}\hfill\smallskip}%
    {\ifthenelse{\equal{#1}{C}}%
      {\item []\hspace*{-\leftmargin}%
       \emph{Greek}\hfill\smallskip}%
      {\ifthenelse{\equal{#1}{D}}%
        {\item []\hspace*{-\leftmargin}%
         \emph{Subscript}\hfill\smallskip}%
          {\ifthenelse{\equal{#1}{E}}%
            {\item []\hspace*{-\leftmargin}%
             \emph{Superscript}\hfill\smallskip}%
              {\ifthenelse{\equal{#1}{F}}%
          {\item []\hspace*{-\leftmargin}%
           \emph{Conventions}\hfill\smallskip}%
              {}%
          }%
        }%
      }%
    }%
  }

\usepackage{tikz}   
\usepackage{arxiv}
\usepackage{caption}
\usepackage[table,xcdraw]{xcolor}
\usepackage{listings}
\usepackage{color}
\usepackage[utf8]{inputenc} 
\usepackage[T1]{fontenc}    
\usepackage{hyperref}       
\usepackage{url}            
\usepackage{booktabs}       
\usepackage{amsfonts}       
\usepackage{nicefrac}       
\usepackage{microtype}      
\usepackage{lipsum}
\usepackage{graphicx}
\usepackage{float}
\graphicspath{ {./images/} }
\lstdefinestyle{myTerminalStyle}{
    basicstyle=\ttfamily\small,
    backgroundcolor=\color{gray!10},
    frame=single,
    framesep=5pt,
    rulecolor=\color{black!60},
    breaklines=true,
    breakatwhitespace=true,
    captionpos=b,
    showstringspaces=false,
    numbers=none
}
\hypersetup{
  colorlinks=true,
  anchorcolor=black,
  citecolor=blue,
  filecolor=black,
  urlcolor=black,
  linkcolor=black
}

\title{arcjetCV: an open-source software to analyze material ablation}

\author{
    Alexandre Quintart \\
  Flying Squirrel\\
  Bourg-Saint-Pierre, Valais 1946, Switzerland \\
  \texttt{alex@flying-squirrel.space} \\
  \And
 Magnus Haw \\
  Thermal Protection Materials Branch\\
  NASA Ames Research Center\\
  Moffett Field, CA 94035, USA \\
  \texttt{magnus.haw@nasa.gov} \\
   \And
 Federico Semeraro \\
  Thermal Protection Materials Branch\\
  AMA at NASA Ames Research Center\\
  Moffett Field, CA 94035, USA \\
  \texttt{federico.semeraro@nasa.gov} \\
}

\begin{document}
\maketitle
\begin{abstract}

arcjetCV is an open-source Python software designed to automate time-resolved measurements of heatshield material recession and recession rates from arcjet test video footage. This new automated and accessible capability greatly exceeds previous manual extraction methods, enabling rapid and detailed characterization of material recession for any sample with a profile video. arcjetCV automates the video segmentation process using machine learning models, including a one-dimensional (1D) Convolutional Neural Network (CNN) to infer the time-window of interest, a two-dimensional (2D) CNN for image and edge segmentation, and a Local Outlier Factor (LOF) for outlier filtering. A graphical user interface (GUI) simplifies the user experience and an application programming interface (API) allows users to call the core functions from scripts, enabling video batch processing. arcjetCV's capability to measure time-resolved recession in turn enables characterization of non-linear processes (shrinkage, swelling, melt flows, etc.), contributing to higher fidelity validation and improved modeling of heatshield material performance. The source code associated with this article can be found at \href{https://github.com/magnus-haw/arcjetCV}{https://github.com/magnus-haw/arcjetCV}. \\

\textbf{Keywords}: Arcjet; Recession Tracking; Machine Learning; Computer vision; Porous Ablator; Open-source.
\end{abstract}


\section{Introduction}
Arcjet Computer Vision (arcjetCV) is a software application built to automate the analysis of arcjet video footage obtained during ground tests of Thermal Protection System (TPS) materials. This includes tracking material recession, sting arm motion, and the shock-material standoff distance. Arcjets are plasma wind tunnels used to test the performance of heatshield materials for spacecraft atmospheric entry. These facilities present a high-enthalpy flow environment with heat fluxes from 5 to 6000 kW/m$^2$ large stagnation pressures ($>$10 kPa) and partially ionized ($\sim$2 eV) plasma similar to atmospheric entry conditions. 

Historically, material samples are only measured before and after an arcjet test to characterize the total recession. These pre/post-test measurements do not capture time-dependent effects such as material expansion, swelling, and non-linear recession. However, arcjet video footage does contain all of this information and is already a standard diagnostic for most tests. Consequently, a wealth of material data can be extracted from video footage. This, at minimum, involves locating the leading edge of the material sample in a video at each frame and tracking the motion of that edge as the test progresses. Due to the highly variable video lighting, sample color, camera focus, background features, flow color, and sample motion, this task is onerous to complete manually for multiple videos. As a result, very few time-resolved measurements of arcjet material recession, expansion, or shrinkage exist. 

arcjetCV was specifically developed to resolve this issue and uses computer vision and machine learning methods to automate the video processing so that time-resolved material behavior can be easily extracted for every arcjet test. Applications of arcjetCV have revealed several unexpected non-linear effects in addition to providing critical recession rate values. In addition, the sample bowshock is also tracked providing additional validation metrics (e.g., time-resolved shock standoff distance, shock shape).

\section{Software Architecture}

arcjetCV is developed in Python and using PySide6 in conjunction with matplotlib \cite{Hunter:2007} to construct its Graphical User Interface (GUI). At the core of arcjetCV's design is the Model-View (MV) architecture, a paradigm that effectively separates the application's data management (the model) from the user interface (the view), enhancing modularity and maintainability.

The model layer plays a critical role in managing the application's central logic and data. It employs OpenCV \cite{opencv_library}, a comprehensive open-source library renowned for its wide range of image processing capabilities, including advanced filtering, transformation, and object detection functions. Additionally, arcjetCV integrates PyTorch \cite{Paszke_PyTorch_An_Imperative_2019} for its machine learning operations, specifically the implementation of CNNs. This modern deep learning library is selected for its efficiency, flexibility, and extensive support for neural network architectures, making it indispensable for arcjetCV's image analysis tasks.

arcjetCV's user interface utilizes PySide6, which provides a vast set of Python bindings to the Qt application framework. This choice ensures a user-friendly and responsive graphical user interface (GUI), allowing users to adjust settings, execute analyses, and review results. The GUI's design is geared towards facilitating an intuitive user experience, ensuring that complex image processing and analysis tasks are accessible to a wide range of researchers.

\subsection{Graphical User Interface}
\begin{figure}[H]
 \centering
 \includegraphics[width=\textwidth]{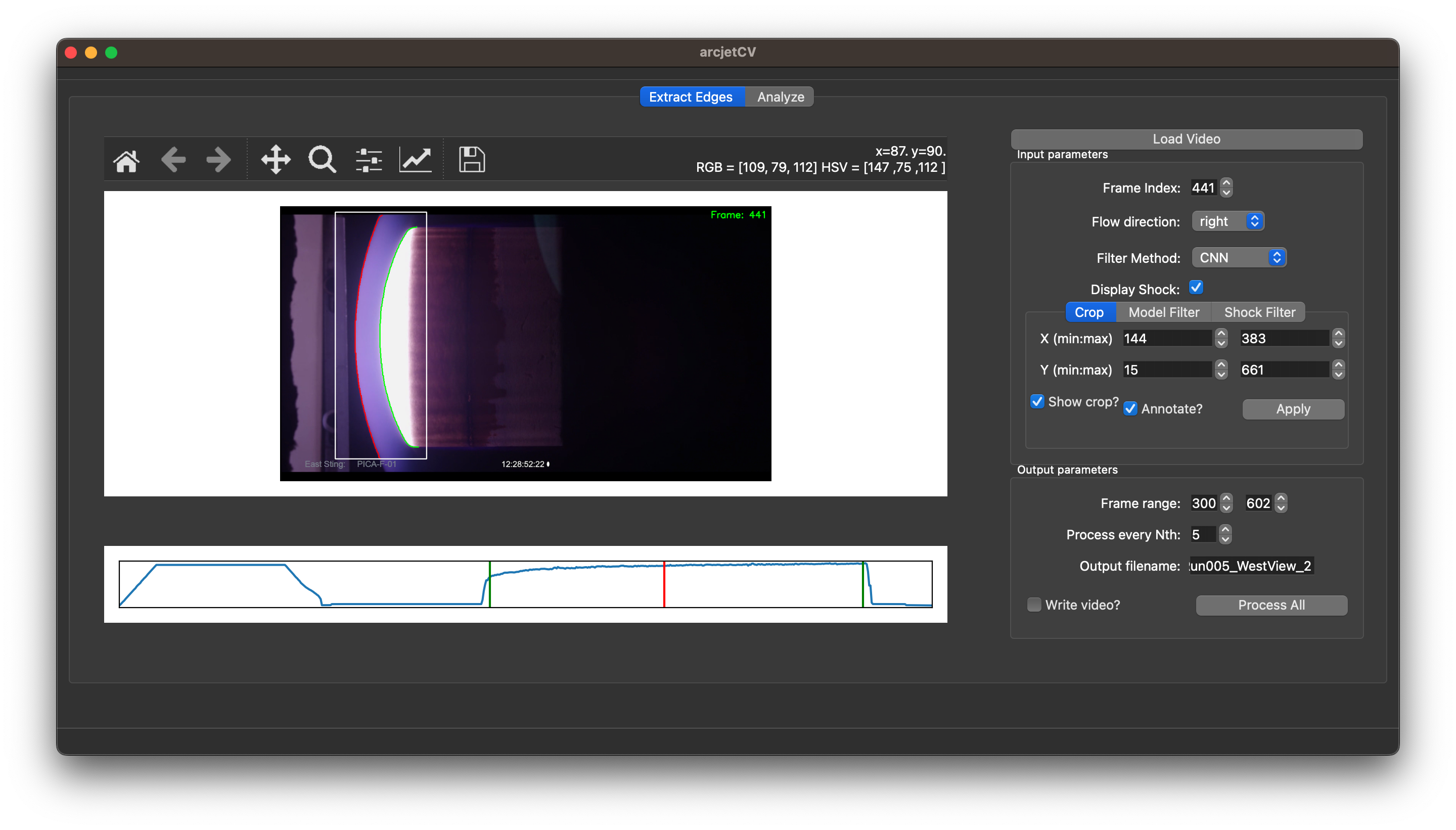}
 \caption{GUI Video Processing window}
 \label{fig:gui1}
\end{figure}

The GUI consists of two main tabs corresponding to the primary stages of analysis provided by arcjetCV: "Extract Edges" for video processing (figure \ref{fig:gui1}), and "Analysis" for post-processing the results (figure \ref{fig:gui2}).

The video processing tab is arranged to identify and extract the edges of the sample and/or shock from video frames. The profiles obtained from this extraction are saved in a JSON file for subsequent analysis. The post-processing tab allows the visualization of these extracted edges and further refines the data to plot material recession over time. Notably, arcjetCV supports several video formats, including *.mp4, *.avi, *.mov, and *.m4v, ensuring wide applicability across various datasets. 

 \subsubsection{Video Processing}

The video processing step is computationally demanding, potentially taking several minutes to finish, depending on the size of the video. To efficiently capture all relevant recession effects without unnecessary processing, analyzing every 10th frame is sufficient for the majority of videos as material recession is typically slow compared to the frame rate. The essential steps in the video processing workflow are as follows:

\begin{enumerate}
    \item Time segmentation of the video (find first and last frames of interest).
    \item Define a region of interest (ROI) to minimize processing per frame.
    \item Identify leading sides of frame (e.g., direction of arcjet flow).
    \item For each frame, classify ROI's pixels into 4 classes, as described in section \ref{sec:imgseg}.
    \item For each frame, extract leading edge from pixels classified as part of the sample.
\end{enumerate}

For the segmentation process to function correctly, the pipeline requires various inputs, including the first and last frames, ROI, and the flow direction. arcjetCV automatically deduces these parameters upon loading a video. The selection of the first and last frames is guided by the time-segmentation model highlighted in section \ref{sec:timeseg}. The ROI determination involves classifying the initial and final frames using the CNN model outlined in section \ref{sec:imgseg}, followed by creating a buffered bounding box around the pixels that do not match the background. The flow direction is determined by analyzing the brighter area in a cumulative image formed by superimposing frames within the specified interest range. Users also have the option to manually adjust these parameters through the GUI or API as needed.

The software offers four segmentation methods for the shock and the sample. An overview of these available segmentation methods is provided below:

\begin{itemize}
    \item \textbf{CNN}: utilizes a convolutional neural network that has been trained on various frames as the core method. While this approach shows strong performance in detecting the edges of shock and sample, it may not perform well on highly underexposed frames.
    \item \textbf{AutoHSV}: employs a combination of preset HSV (Hue, Saturation, Value) ranges to identify contours, opting for the HSV color model over RGB for its simpler and more intuitive thresholding capabilities. This method is  effective for segmenting based on color, saturation, and intensity.
    \item \textbf{HSV}: identifies contours through the use of adjustable HSV ranges. A desktop interface offers real-time display of RGB, HSV, and pixel position values upon mouse hover, aiding the precise adjustment of these ranges. This is especially helpful for manual processing of frames with low exposure.
    \item \textbf{Gray}: detects the leading edge of the sample using a grayscale intensity threshold. This method will only capture a single edge where the other methods will try to capture both sample and shock edges. 
\end{itemize}

Upon configuring all the necessary input parameters, the targeted video sub-sequence can be processed by clicking the "Process All" button. This action generates a JSON file with the intermediate edge data, which is then forwarded to the data analysis tab of the GUI, along with an optional video output. A Python API to handle these requests has also been developed (see section \ref{sec:api}).

The problem of the first frame being underexposed is frequent, causing the automated CNN process to inaccurately detect the edge. In situations where the recession significantly outpaces the frame rate, this issue can be addressed by applying one of the manual filter options to the initial frame of interest and then exporting the outcome. The post-processing phase is designed to accommodate multiple processed sequences, facilitating edge case corrections.

\subsubsection{Data Analysis}

\begin{figure}[H]
 \centering
 \includegraphics[width=\textwidth]{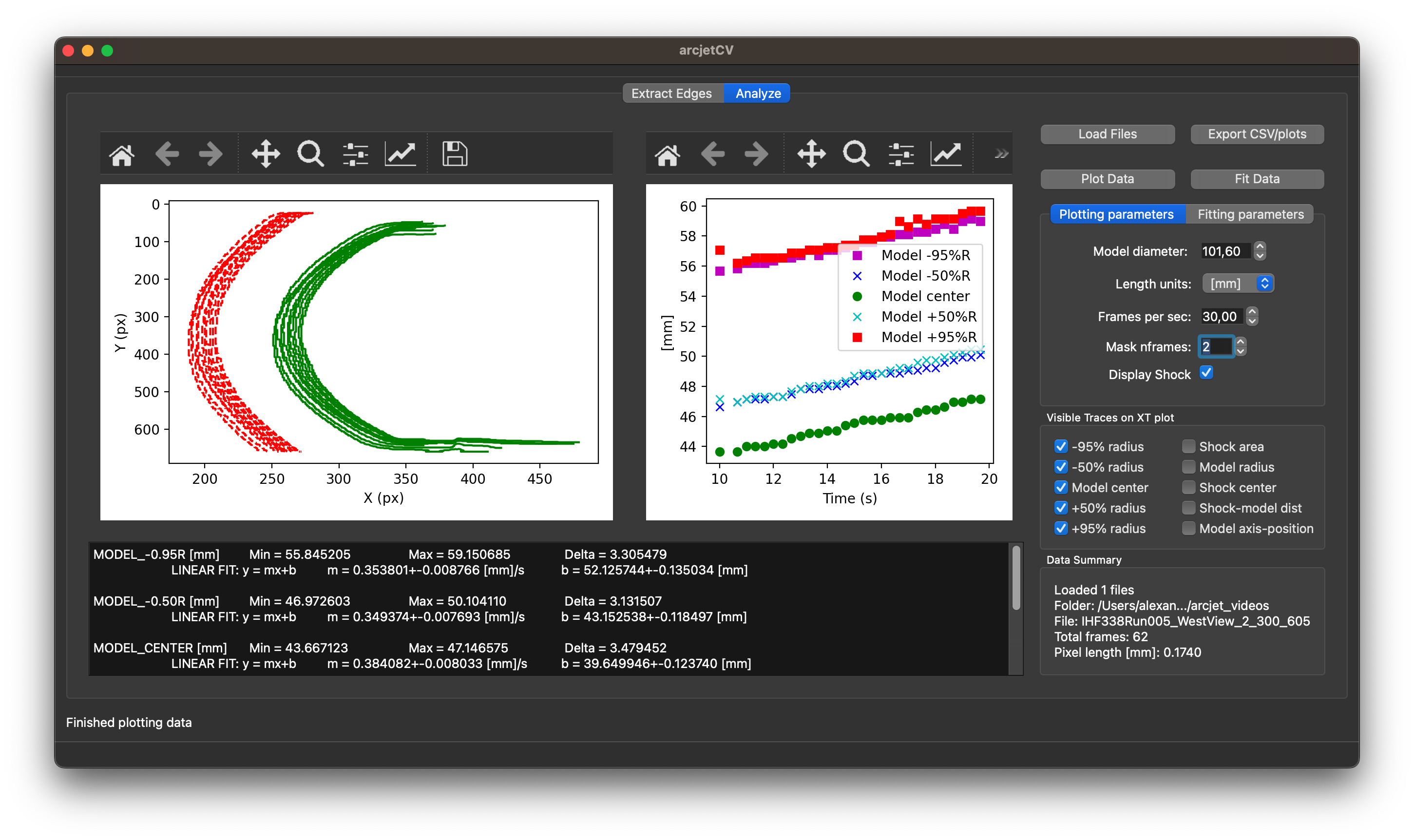}
 \caption{GUI Data Analysis window}
 \label{fig:gui2}
\end{figure}

The post-processing stage in the pipeline focuses on plotting, cleaning, calibrating, and fitting the edge traces extracted during the video processing phase. The  "Analysis" tab of the GUI (Figure \ref{fig:gui2}) loads a JSON file for a processed video. These loaded edges can then be visualized with the "Plot Data" button which will display the traces on the two plotting windows. The left plot (XY) shows a subset of the extracted traces and the right plot (TX) displays a range of possible time series data (e.g, recession at discrete positions as function of sample radius, sample area, shock-sample standoff distance, model vertical position, etc.). The pixel values are also scaled to physical units based on the model diameter.

Upon displaying the data, users have the option to apply linear fits to the visible time series, with both the fits and corresponding fit parameter error bars showcased in the lower part of the window. This feature is particularly valuable for analyzing test cases characterized by linear recession, facilitating the standard processing workflow. However, a poor linear fit can also indicate when a particular sample has non-linear recession. To conclude the analysis, the plots, time-series and fitting data can be exported to PNG/CSV files.

\subsection{Python API}\label{sec:api}

The arcjetCV framework is designed to streamline video processing tasks, particularly focusing on the analysis of arcjet videos. It encapsulates this functionality within three primary classes: \texttt{Video}, \texttt{VideoMeta}, and \texttt{ArcjetProcessor}, each serving a distinct role in the video processing pipeline.

\paragraph{Video Class} The foundational layer for manipulating video files, utilizing OpenCV to offer a user-friendly interface for various video operations. It simplifies the complexities of video file handling, providing straightforward access to individual frames, video attributes, and the generation of processed output videos. The class enables easy retrieval of specific video frames for processing, and supports the creation of new video files from processed frames.

\paragraph{VideoMeta Class} Complementing the \texttt{Video} class, \texttt{VideoMeta} specializes in the management of video metadata. It extends Python's dictionary class with JSON-based metadata storage, improving the preservation and accessibility of video processing parameters and analytical data. This class is designed for efficient metadata handling, capturing processing parameters and video analysis results, supporting JSON format for easy metadata saving and retrieval, and allowing for dynamic adjustments to video processing parameters.

\paragraph{ArcjetProcessor Class} At the core of arcjetCV's processing capabilities, this class conducts several image processing tasks, including segmentation, edge detection, and feature extraction. Many of its methods are described next.

\section{Methodology}

In this section, we delve into the methodologies used by arcjetCV to automate video analysis through machine learning. This encompasses image processing techniques as well as post-processing methods designed to consolidate the extracted features into a user-friendly format. By doing so, it enables users to easily derive meaningful insights from the processed video data.

\subsection{Time Segmentation}\label{sec:timeseg}

\begin{figure}[H]
    \centering
    \begin{subfigure}[b]{0.49\textwidth}
         \centering
         \includegraphics[width=0.9\textwidth]{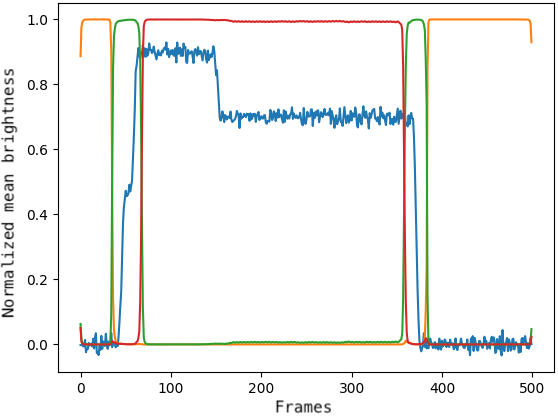}
         \caption{Synthetic signal and labels}
         \label{fig:1dcnn_synth}
    \end{subfigure}
    \hfill
    \begin{subfigure}[b]{0.49\textwidth}
         \centering
         \includegraphics[width=0.9\textwidth]{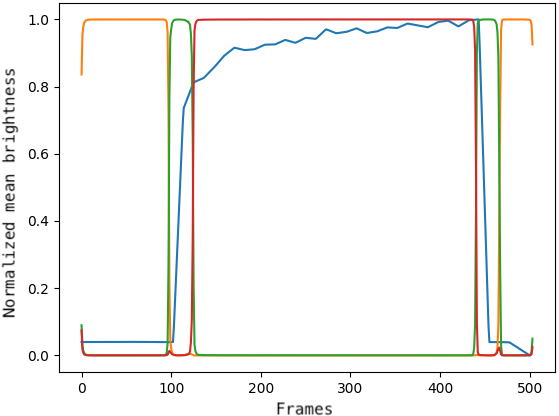}
         \caption{Prediction on real video}
         \label{fig:1dcnn_pred}
    \end{subfigure}
    \caption{Normalized time series derived from the brightness intensity of video frames \label{fig:1dcnn}}
\end{figure}

The process of identifying specific time intervals of interest is automated through a 1D CNN, which has been trained on synthetic brightness intensity data. As illustrated in figure \ref{fig:1dcnn}, our model processes a normalized time series derived from the video frames brightness (in blue), categorizing the data into three distinct classes: \textit{OFF} (in orange), \textit{TRANSITION} (in green), and \textit{ON} (in red). By doing so, the model automates an important step that previously required manual intervention in the analysis workflow.

\paragraph{Architecture}
The Conv1DNet architecture, combines 1D convolutional and transposed convolutional layers. This design effectively captures and reconstructs temporal patterns, crucial for accurate segmentation. The network starts with two convolutional layers, expanding and then compressing the feature space from 1 to 32 and then to 16 channels, respectively. Each layer uses a kernel size of 7, stride of 2, and padding of 3, with 25\% dropout rate after each layer to mitigate overfitting. Following convolution, three transposed convolutional layers incrementally reconstruct the time series dimensionality. These layers adjust the feature space from 16 channels back to 3, correlating to the desired segmentation outputs. ReLU activation functions are applied post each convolutional layer to introduce non-linearity, except for the final layer where a softmax function is utilized. The softmax output enables a probabilistic interpretation of the segmentation results, categorizing each time step into one of the three segmentation classes.

\paragraph{Dataset Generation}
In response to the challenge of the limited amount of training videos available, a synthetic database made of square signals augmented with noise was generated. Additionally, each signal includes an initial spike and a subsequent step function, designed to reflect the dynamic changes typical in natural signals. To further enhance the realism and applicability of these artificial signals, a smoothing process is applied, ensuring a natural integration of noise and signal features. Figure \ref{fig:1dcnn_synth} provides a visual representation of an artificial brightness trace as well as the labeled classes employed to train the time segmentation model.

\begin{figure}[H]
    \centering
    \begin{subfigure}[b]{0.49\textwidth}
         \centering
          \includegraphics[width=\textwidth]{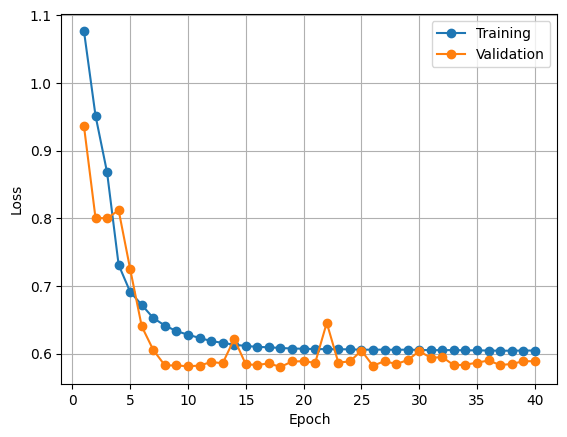}
         \caption{Loss}
         \label{fig:1dcnn_loss}
    \end{subfigure}
    \hfill
    \begin{subfigure}[b]{0.49\textwidth}
         \centering
         \includegraphics[width=\textwidth]{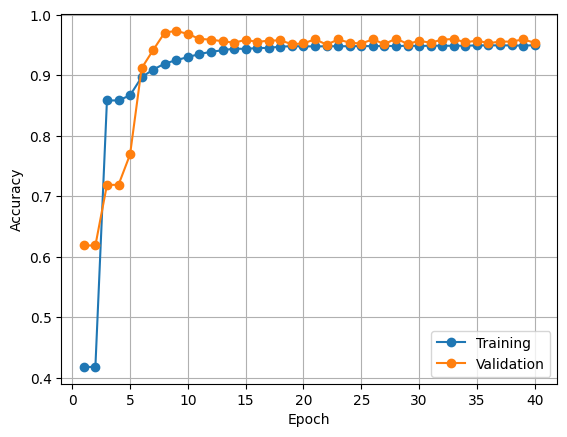}
         \caption{Accuracy}
         \label{fig:1dcnn_acc}
    \end{subfigure}
    \caption{1D CNN model training}
\end{figure}

\paragraph{Training and Results}
The training was conducted over 40 epochs, as shown in figure \ref{fig:1dcnn_loss}, using a dataset of 1000 synthetic samples. This extended training period was necessary for the model to iteratively learn and adjust its weights, where each epoch represented a full pass through the dataset. The choice of 1000 samples was observed to offer a sufficiently varied set of data for effective training, while also keeping a reasonable computational requirements. Figure \ref{fig:1dcnn_pred} illustrates the 1D CNN accurately predicting phases in a video's normalized brightness signal. At the end of the training the 1D CNN reached an accuracy of 98\% on the validation dataset, as shown in figure \ref{fig:1dcnn_acc}.

\subsection{Local Outlier Factor}
In developing arcjetCV, the challenge of video frame quality, such as saturation and blur, is addressed by utilizing the LOF method from the scikit-learn \cite{scikit-learn} library. This technique identifies and filters out anomalous frames by assessing their deviation from the local density of their neighbors, effectively pinpointing frames that are outliers due to unusual characteristics. This approach proves particularly effective for arcjetCV, since arcjet video frame quality can vary drastically, ensuring that only high-quality frames are processed.

\subsection{Image Segmentation and Edge Detection}\label{sec:imgseg}

\begin{figure}[H]
\centering
 \centering
 \includegraphics[width=\linewidth]{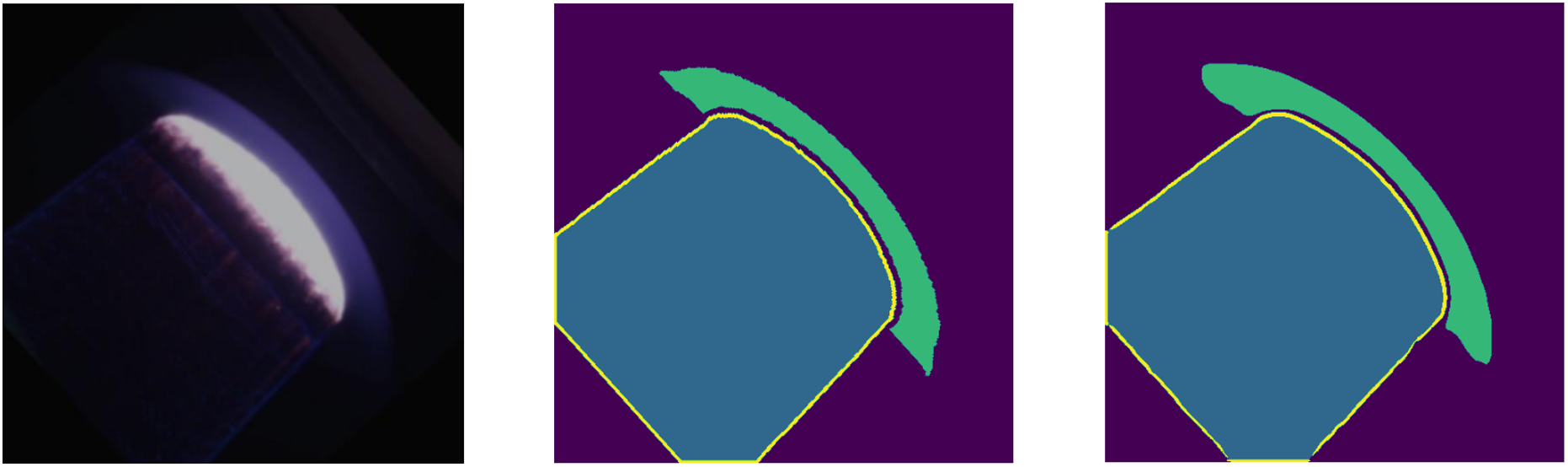}
     \caption{Challenging video frame (left), ground truth masks (middle), and Xception prediction (right)}
     \label{fig:predictions}
\end{figure}

A 2D CNN was designed to improve the automatic segmentation of videos, aiming to track the leading edges of samples and shocks. The model predicts 4 classes: \textit{background}, \textit{sample}, \textit{sample-edge} and \textit{shock} (see figure \ref{fig:predictions}). Pretrained on ImageNet \cite{deng2009imagenet}, the model was fine-tuned on a manually annotated datasets described below. The trained network demonstrated strong performance, substantially decreasing the requirement for manual intervention across all tested videos.

\paragraph{Architecture}
The model combines a standard UNet architecture with the Xception encoder \cite{chollet2017xception} from the "Segmentation Models Pytorch" package \cite{Iakubovskii:2019}. The frame of the video to be analyzed is first scaled with padding to a 256x256 square, which is then passed to the model for prediction. Structured into three main flows, the model features an entry flow for initial feature extraction, a middle flow dedicated to deeper analysis through the repeated application of depthwise separable convolutions, and an exit flow that compiles and prepares the outputs. Choosing a UNet followed a comparative analysis and experimentation with various architectures, including FCN, SegNet, PSPNet. Its ability to capture fine details and its original development for medical imaging, a field requiring precision similar to that needed for analyzing arcjet frames, were key factors in this decision. On the other hand, the choice of Xception as the underlying architecture is motivated by its balance between computational efficiency and robust performance, making it advantageous for rapid training and achieving excellent results with limited data inputs.

\paragraph{Dataset Generation}
For the training of the machine learning models within arcjetCV, it was imperative to construct a comprehensive dataset. This dataset was  assembled from 400 frames extracted from 40 distinct arcjet videos. Each of these frames underwent a manual segmentation process, resulting in the creation of masks that delineate the four classes of interest. To streamline the segmentation process of our dataset, we developed a specialized tool based on the Segment Anything Model (SAM) \cite{kirillov2023segment,samgui,semeraro2023tomosam}, combined with the grabcut function \cite{rother2004grabcut} provided by OpenCV. Following the segmentation of the original frames, the dataset was extended using a series of data augmentation techniques provided by the Albumentations library \cite{info11020125} to enhance its diversity. These augmentation strategies included rotating the frames, altering the positions of the segmented elements within the frames, and modifying the color schemes, amounting to a final 1778 training frames. 

\begin{figure}[H]
    \centering
    \begin{subfigure}[b]{0.33\textwidth}
        \centering
        \includegraphics[width=\textwidth]{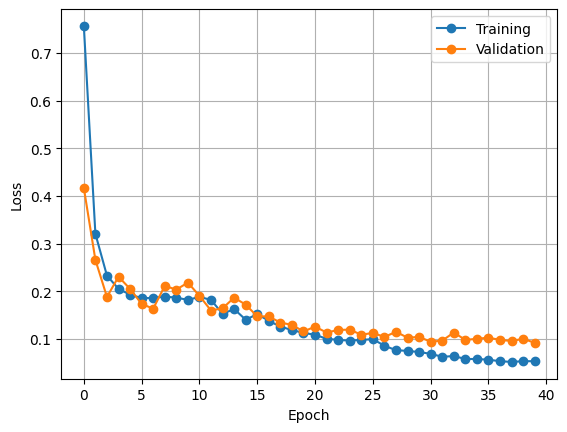}
        \caption{Loss}
        \label{fig:loss}
    \end{subfigure}
    \hfill
    \begin{subfigure}[b]{0.33\textwidth}
        \centering
        \includegraphics[width=\textwidth]{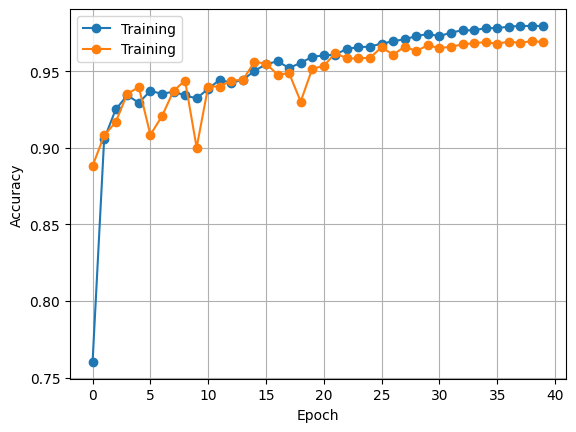}
        \caption{Accuracy}
        \label{fig:acc}
    \end{subfigure}
    \hfill
    \begin{subfigure}[b]{0.33\textwidth}
        \centering
        \includegraphics[width=\textwidth]{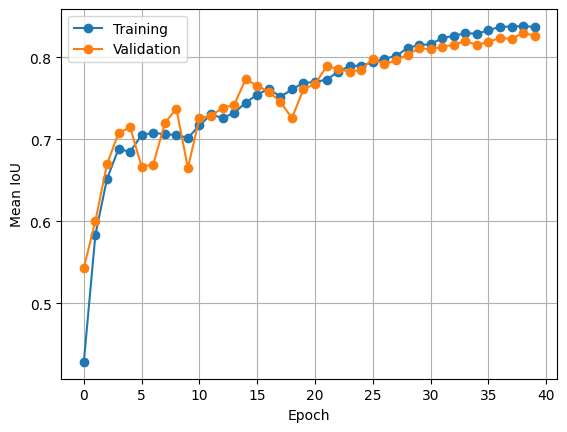}
        \caption{Mean IoU}
        \label{fig:score}
    \end{subfigure}
    \caption{Metrics for the Xception model trained up to 40 epochs}
\end{figure}

\paragraph{Training and Results}
Three distinct UNet models were considered, as shown in table \ref{tab:model_comparison}, with training durations ranging from 15 to 40 epochs. The Xception model stood out for its superior accuracy performance and the 25-epoch training regimen was chosen to avoid overfitting (see figure \ref{fig:acc}). As indicated in tables \ref{tab:model_comparison} and \ref{tab:epochs_accuracy_miou}, the Xception model delivers the highest accuracy by pixel and mean Intersection over Union (mIoU) on the test set, as compared to ResNet18 and VGG16, which were predominantly used during past iterations of arcjetCV \cite{quintart2023arcjetCV}. In addition, figure \ref{fig:predictions} showcases a side-by-side view of the original video frame, the ground truth mask, and the prediction produced by the Xception model after it was trained for 25 epochs.

\noindent
\begin{minipage}[t]{0.45\textwidth}
\centering
\captionof{table}{Comparison of best Pixel Accuracy and mIoU across different models \label{tab:model_comparison}}
\begin{tabular}{|c|c|c|c|}
\hline
\rowcolor[HTML]{C0C0C0} 
\textbf{Model} & \multicolumn{1}{c|}{\textbf{VGG16}} & \multicolumn{1}{c|}{\textbf{ResNet18}} & \multicolumn{1}{c|}{\textbf{Xception}} \\ \hline
\cellcolor[HTML]{C0C0C0}\textbf{Pixel Accuracy} & 0.9411 & 0.9413 & 0.9687 \\ \hline
\cellcolor[HTML]{C0C0C0}\textbf{mIoU} & 0.7311 & 0.7543 & 0.8005 \\ \hline
\end{tabular}
\end{minipage}
\hfill
\begin{minipage}[t]{0.45\textwidth}
\centering
\captionof{table}{Comparison of Pixel Accuracy and mIoU for the Xception model across different epochs \label{tab:epochs_accuracy_miou}}
\begin{tabular}{|c|c|c|}
\hline
\rowcolor[HTML]{C0C0C0} 
\textbf{Epochs} & \textbf{Pixel Accuracy} & \textbf{mIoU} \\ \hline
\cellcolor[HTML]{C0C0C0}15 & 0.9482 & 0.7618\\ \hline
\cellcolor[HTML]{C0C0C0}20 & 0.9645 & 0.7897\\ \hline
\cellcolor[HTML]{C0C0C0}25 & 0.9687 & 0.8005 \\ \hline
\cellcolor[HTML]{C0C0C0}30 & 0.9657 & 0.7993 \\ \hline
\cellcolor[HTML]{C0C0C0}40 & 0.9662 & 0.8097 \\ \hline
\end{tabular}
\end{minipage}

\section{Enabling New Analysis and Validation Capabilities}
The fundamental goal of arcjetCV is to enable higher fidelity analysis and validation for arcjet testing. This section describes some of the new analyses and validation capabilities enabled by arcjetCV.

\subsection{Non-linear Recession}
Although most heatshield materials have linear ablation under nominal conditions, all materials will exhibit non-linear recession under certain conditions (e.g., non-linear test conditions, testing to failure, swelling, shrinkage). For these circumstances, acquisition of time-resolved recession is essential for analysis of the test and the material performance. 

\begin{figure}[H]
    \centering
    \includegraphics[width=0.6\textwidth]{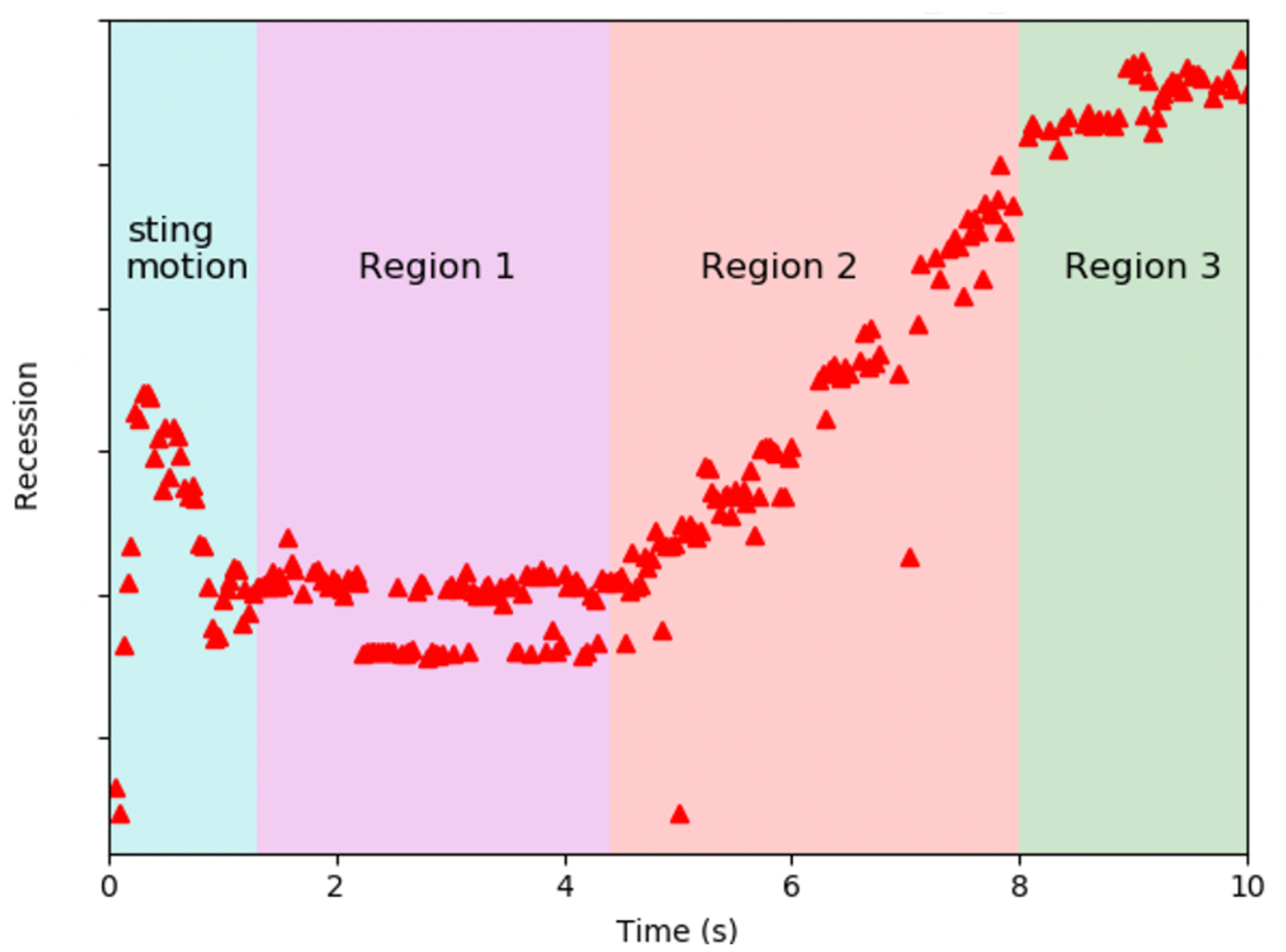}
    \caption{Centerline tracking of material leading-edge showing non-linear recession}
    \label{fig:nonline}
\end{figure}

One simple example of this is NASA's Heat shield for Extreme Entry Environment Technology (HEEET) material \cite{tavares2020ethical}, a 3D woven composite. This material has two distinct layers, a high density ablation layer and a low-density insulating layer that are woven together. Recovering time-resolved recession is necessary to capture the different layer behaviors. The non-linear recession shown in Figure \ref{fig:nonline}, highlights three distinct recession rates in a constant aerothermal environment. After the sting arm stabilizes, the first seconds show little to no recession (region 1). The second region has a nearly constant recession rate, and the third region exhibits a markedly reduced recession rate. 

Recent testing of PICA and Avcoat materials also indicate a critical need for material models which can account for non-linear effects such as material swelling and shrinkage at the microscale \cite{ferguson2021update, toolkit2023}. These models can only be validated by in-situ tracking during high-temperature tests, which will be a future application of arcjetCV.

\subsection{Shape Change}

\begin{figure}[H]
    \centering
    \begin{subfigure}[b]{0.64\textwidth}
        \centering
        \includegraphics[width=\textwidth]{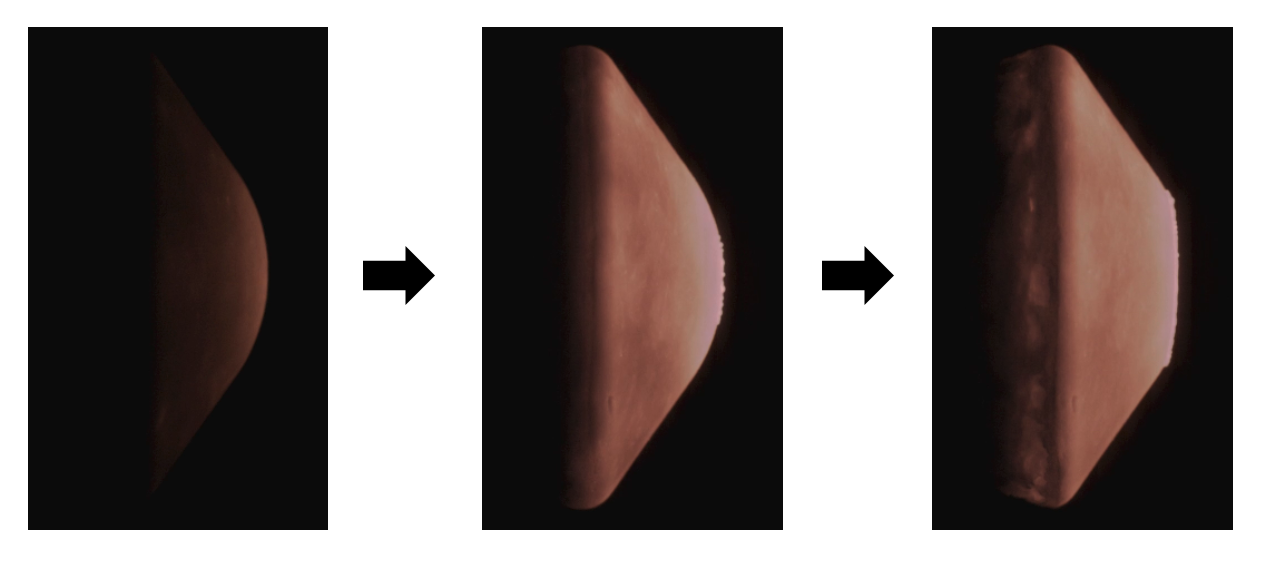}
        \caption{Frames from arcjet video}
    \end{subfigure}
    \hfill
    \begin{subfigure}[b]{0.35\textwidth}
         \centering
         \includegraphics[width=.95\linewidth]{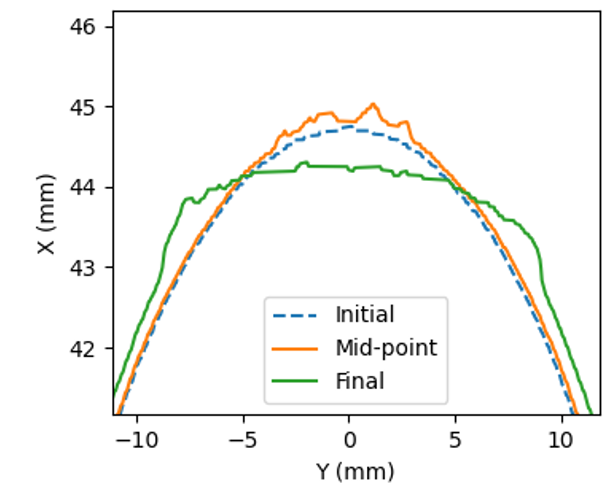}
         \caption{Shape-change over time}
    \end{subfigure}
    \caption{Non-linear recession with significant shape change for a PICA-NuSil sample at HyMETS}
    \label{fig:shape_change}
\end{figure}

arcjetCV analysis also enables characterization of the changing shape of various samples during tests. As illustrated in figure \ref{fig:shape_change}, the plots show the initial and final contours for a PICA-NuSil sample from a test conducted at NASA Langley Research Center's HyMETS arcjet. The high-fidelity segmentation of the images, achieved with a resolution of 0.1 mm, reveals several notable effects. Firstly, the sample's nose experiences flattening, along with the formation of millimeter-sized ripples or bubbles. Secondly, there's a noticeable overall expansion of the sample, as highlighted by the negative recession at the edges. This example shows several new forms of investigation enabled by arcjetCV: analysis of bulk shape change, tracking melt flows, and sample swelling.

\subsection{Shock Standoff Validation}
The shock standoff distance is another valuable form of validation that arcjetCV analysis provides. This distance is a simple validation metric for comparison with computational fluid dynamics (CFD) simulations that are used to interpret the results of arcjet testing as well as scale them to flight. Confirming the standoff distance matches the simulated values is a simple sanity check. More involved analysis that simulates shape change can also be verified by checking the changing shock standoff distance. Furthermore, for in-situ diagnostics including stagnation pressure sensors and calorimeters, confirming the standoff distance is as expected is an excellent metric to detect undesired pressure leaks: calibration models with a smaller standoff than expected will typically have a pressure leak that needs to be repaired.

\begin{figure}[H]
    \centering
    \begin{subfigure}[b]{0.56\textwidth}
        \centering
        \includegraphics[width=\textwidth]{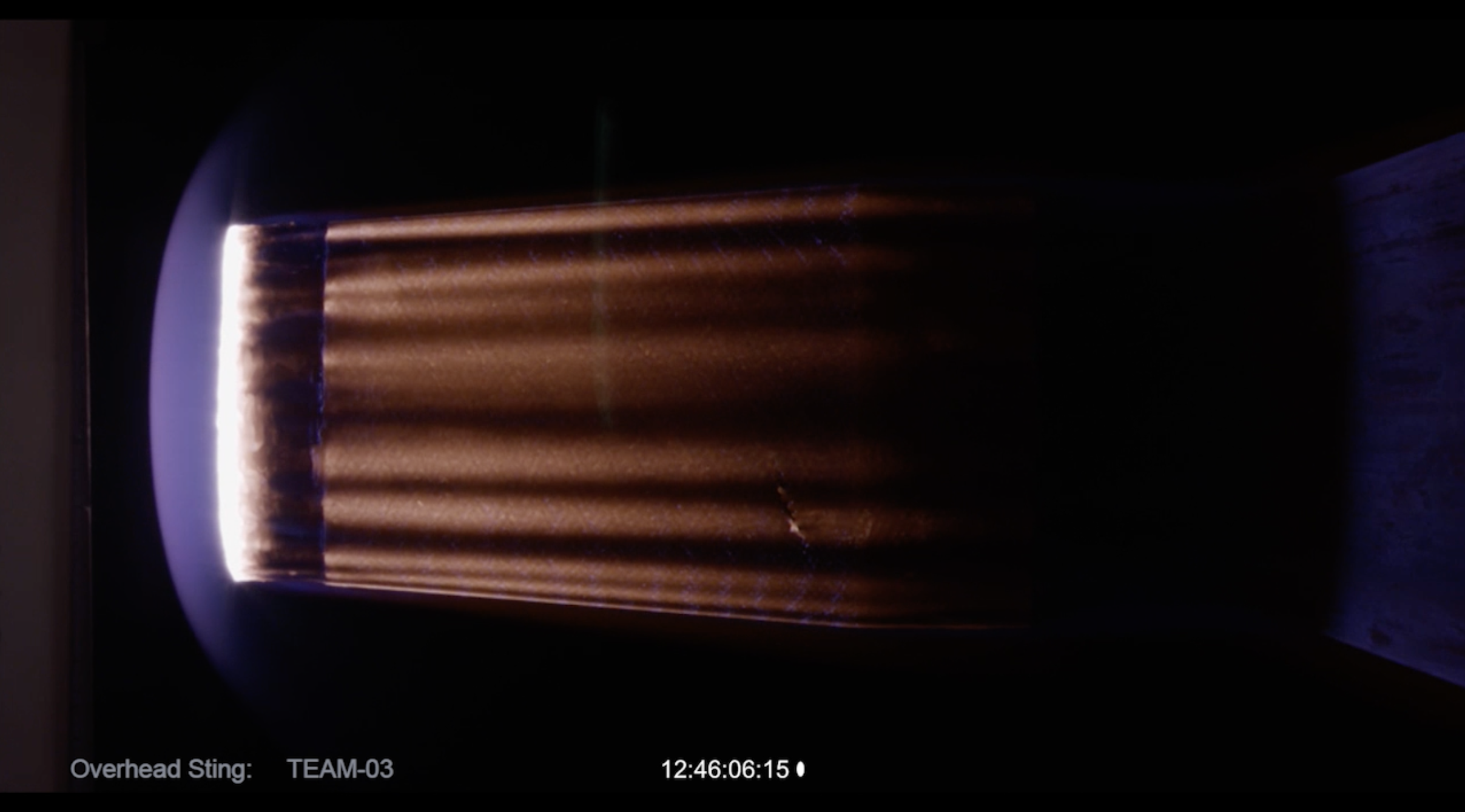}
        \caption{Frame from arcjet video with a flat face sample}
    \end{subfigure}
    \hfill
    \begin{subfigure}[b]{0.43\textwidth}
         \centering
         \includegraphics[width=\textwidth]{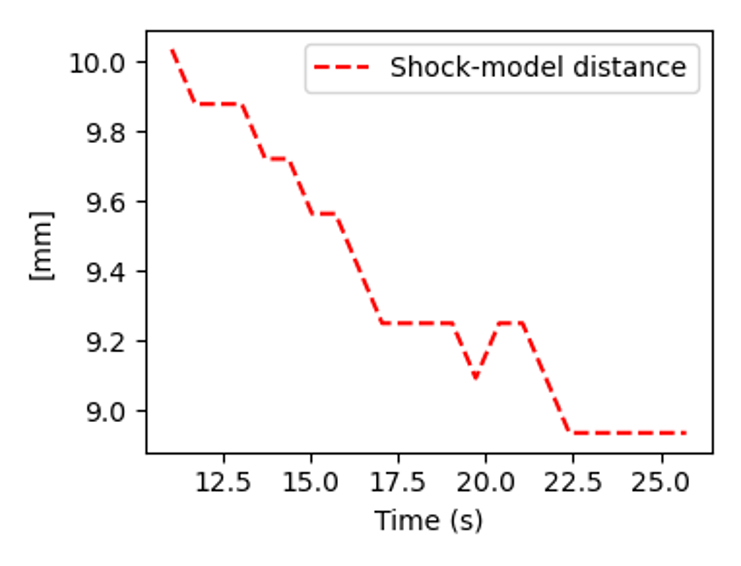}
    \caption{Plot of sample shock-standoff distance over time}
    \label{fig:shockstand}
    \end{subfigure}
    \caption{Shock standoff distance over time for an arcjet test with a flat face sample}
    \label{fig:shock}
\end{figure}

\section{Conclusion}
In this paper, we introduced the open-source software arcjetCV, an innovative tool designed to segment and analyze videos of arcjet tests by leveraging machine learning techniques. Through rigorous testing and evaluation, arcjetCV has demonstrated good performance in terms of precision, processing speed, and adaptability, showcasing its effectiveness in handling the dynamic and complex phenomena observed during arcjet environments. The application's interface and streamlined processing pipeline render it accessible to a broad spectrum of users, from technicians to scientists. Its automated segmentation and analysis features significantly save time and resources, while delivering consistent results, thus reducing the influence of human error.

The development of arcjetCV is a continuous endeavor, with many opportunities for further enhancements and fine-tuning. Several of its current limitations are due to inherent challenges in video capture and processing, which impact the accuracy of the software. These include projection errors, camera misalignment, as well as poor exposure and depth of field, complicating shock and sample recognition and tracking. Future work will aim at overcoming existing challenges, such as the analysis of videos with varying degrees of noise or low visibility, as well as 3D reconstruction of the sample's surface through the use of a stereo camera setup. 

 In conclusion, the introduction of modern computer vision techniques for the segmentation and analysis of arcjet videos represents a significant improvement in the TPS testing and modeling domain. Notably, arcjetCV is currently utilized by the NASA Ames Research Center arcjet facilities in support of several missions, including Mars Sample Return (MSR), Dragonfly, and the Artemis lunar program.

\section{Acknowledgements}
This work was supported by internal research and development (IRAD) funds from the Thermal Protection Materials branch of NASA Ames Research Center. The authors would particularly like to thank Mairead Stackpoole, the Mars Sample Return project, and the arcjet team for providing test videos and for their feedback.

\bibliographystyle{unsrt}  
\bibliography{references}

\end{document}